\documentclass[letterpaper]{article}
\usepackage{aaai19}
\usepackage{times}
\usepackage{helvet}
\usepackage{courier}
\usepackage{graphicx}
\usepackage{tabularx}
\usepackage{amsmath}
\usepackage{amssymb}
\usepackage{color}

\begin{document}
\title{Systematic Analysis and Removal of Circular Artifacts for StyleGAN}
\author{Way Tan, Bihan Wen, Xulei Yang}
\maketitle
\maketitle
\begin{abstract}
StyleGAN is one of the state-of-the-art image generators which is well-known for synthesizing high-resolution and hyper-realistic face images. Though images generated by vanilla StyleGAN model are visually appealing, they sometimes contain prominent circular artifacts which severely degrade the quality of generated images. In this work, we provide a systematic investigation on how those circular artifacts are formed by studying the functionalities of different stages of vanilla StyleGAN architecture, with both mechanism analysis and extensive experiments. The key
modules of vanilla StyleGAN that promote such undesired artifacts are highlighted. 
Our investigation also explains why the artifacts are usually circular, relatively small and rarely split into 2 or more parts. Besides, we propose a simple yet effective solution to remove the prominent circular artifacts for vanilla StyleGAN, by applying a novel pixel-instance normalization (PIN) layer.
\end{abstract}

\section{Introduction}
Since generative adversarial networks (GANs) were introduced in 2014 by Google Researcher Ian Goodfellow \cite{GAN_2014}, the technique has been widely adopted in image generation and can now produce highly convincing fake images of anime characters \cite{anime_2017}, landscapes \cite{GauGAN_2019}, human faces \cite{PRGAN_2017}, etc. Especially for facial generation, GANs have made huge breakthroughs in the past few years \cite{GAN_2014,DCGAN_2015,CoupledGAN_2016,PRGAN_2017,BigGan_2018}. Amongst them, StyleGAN \cite{StyleGAN_2018} is the current state-of-the-art image generator for high-resolution and hyper-realistic face images\footnote{The authors primarily demonstrated the applicability of StyleGAN on facial generation, though they also highlighted that StyleGAN can be extended to other objects (e.g. cars, rooms, animals) given appropriate training datasets.}, with size up to $1024 \times 1024$. More importantly, StyleGAN uses a new generator architecture to give control over the disentangled style properties of generated facial images. The quality of generated faces has been evaluated to be superior to most other models by a variety of metrics \cite{Eval_2019,HYPE_2019}. 

\begin{figure}[!t]
    \centering
        \includegraphics[width=2.6cm]{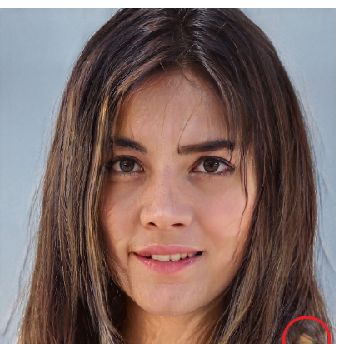}
        \includegraphics[width=2.6cm]{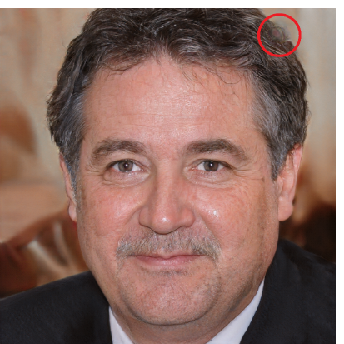}
        \includegraphics[width=2.6cm]{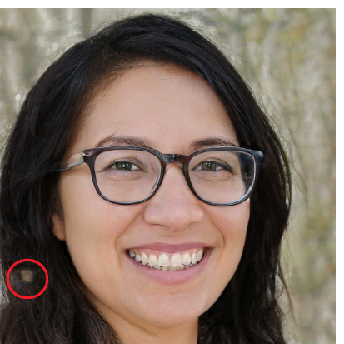}
    \caption{Random examples of the generated face images using the vanilla StyleGAN model. The circular artifacts in the generated examples are highlighted by red circles. }
    \label{fig:Embedart}
    \vspace{-0.1in}
\end{figure}

However, the generated ``natural'' images by StyleGAN are still not ``perfect'' due to the presence of various artifacts. 
For example, circular artifacts are commonly observed in the images generated by StyleGAN. Fig.~\ref{fig:Embedart} shows some examples of the face images with the undesired circular artifacts that are randomly generated by StyleGAN. Such circular artifacts have been widely reported~\cite{Image2Style_2019,style-gan2}.
Amongst these artifacts, circular artifacts are arguably the most severe image degradation.
Our empirical results showed that nearly \textit{all} generated face images by StyleGAN contain such artifacts which degrade the image quality significantly.

In this work, we provide a systematic investigation of the circular artifacts generated by StyleGAN~\cite{StyleGAN_2018}. 
We first characterise the circular artifacts via a case study, followed by analysis on the cause of such artifacts and proposing three claims based on the analysis. 
We provide justifications as well as several empirical evidences to support our claims.
To the best of our knowledge, no systematic study\footnote{The improved version of StyleGAN \cite{style-gan2} attributed the source of water droplets to restrictions on the generator imposed by the Adaptive Instance Normalization (AdaIN), and therefore replaced it with weight demodulation to remove circular artifacts. However, the underlying causes were not yet investigated.} has to date been carried out to investigate the cause of these StyleGAN-generated artifacts.
Based on the analysis, we propose a simple yet effective solution to remove the prominent circular artifacts for vanilla StyleGAN~\cite{StyleGAN_2018}, by applying a novel pixel-instance normalization (PIN) layer.

\section{Analysis of StyleGAN and Generated Artifacts}

\begin{figure}[t]
\centering
\includegraphics[width=8.1cm]{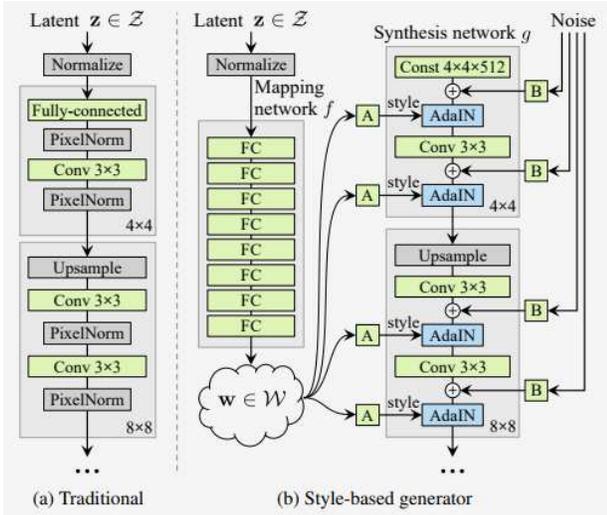}
\caption{The architecture of the vanilla StyleGAN: the illustration taken from the original StyleGAN paper~\cite{StyleGAN_2018}.}
\label{fig:StyleGANarch}
\vspace{-0.1in}
\end{figure}

StyleGAN makes use of the architecture introduced by progressive GAN (PRGAN) \cite{PRGAN_2017}, which can synthesize very large high-quality images by growing both the discriminator and generator models during the training process. 
Figure~\ref{fig:StyleGANarch} (a) shows the architecture of PRGAN, which starts with low-resolution, i.e., $4\times4$ images, and progressively increases to higher resolutions, i.e., $8\times8$, $16\times16$, etc., with 2 layers at each resolution. 
Figure~\ref{fig:StyleGANarch} (b) illustrates the StyleGAN architecture which retains this progressive structure with 2 layers per resolution, but makes a number of changes to the synthesis network architecture.

For one, the generator of StyleGAN takes a learned constant instead of a vector from the latent space as input. To achieve variation in generated images, there are two new sources of randomness: a MLP mapping network and noise inputs after each convolution.

Learned affine transformations are applied to the output from the mapping network, and these values are used to define a style to be transferred via Adaptive Instance Normalization (AdaIN) at each resolution \cite{styleTransfer_2017}.
Further stochastic variation is explicitly introduced by adding uncorrelated per-pixel Gaussian noise inputs, which are then scaled by a learnt scaling factor.


\begin{figure}[!t]
\centering
\includegraphics[width=8.4cm]{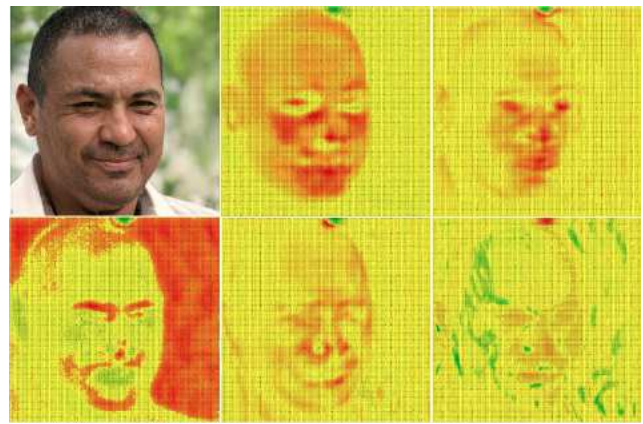}
\caption{Selected feature maps at $128\times128$ resolution after the 1st AdaIN, shown alongside the final generated face (top-left). Green and Red, respectively, denote the positive and negative activations of highest magnitude. The artifact region at the top of the image has a consistently high-magnitude activation in all feature maps.}
\label{fig:128x128fmap}
\vspace{-0.1in}
\end{figure}

\subsection{A Case Study}
Considering an example face image in Fig.~\ref{fig:128x128fmap} generated by the vanilla StyleGAN, there is clearly a circular artifact located at the top of the image. 
To illustrate how such artifacts are synthesized, we visualize the $128\times128$ feature maps, across all channels, from the layer after the first AdaIN of StyleGAN in Fig.~\ref{fig:128x128fmap}. We have the following observations:
\begin{enumerate}
    \item Most of the feature energy (i.e., peak activation zones in red or green) is distributed around objects of interest, dubbed ``face region'', comparing to the background, dubbed ``background region''.
    \item There is a peak spot that consistently recurs across all feature maps, and its location is similar to that of the circular artifact in the generated face image. We refer to such locations in the immediate features as the ``artifact region''.
\end{enumerate}

\begin{figure}[!ht]
\centering
\includegraphics[width=8cm]{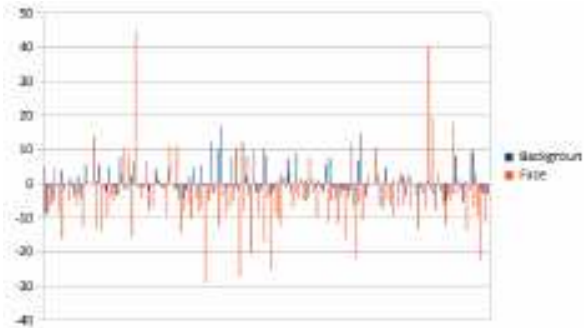} \\
{\small (a) \textit{Background} vs \textit{Face} Regions at $128 \times 128$ feature map.} \\
\includegraphics[width=8cm]{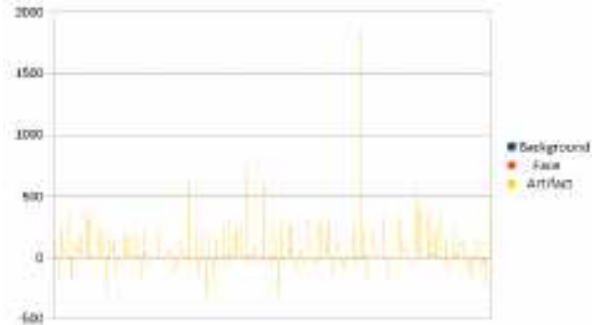} 
{\small (c) \textit{Background} vs \textit{Face} vs \textit{Artifacts} Regions, } \\
{\small at $128 \times 128$ feature maps.} \\
\includegraphics[width=8cm]{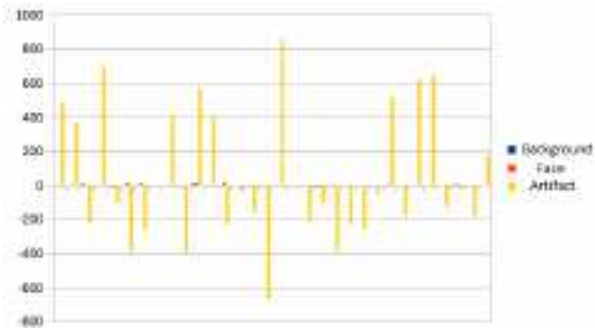}
{\small (c) \textit{Background} vs \textit{Face} vs \textit{Artifacts} Regions, } \\
{\small at higher-resolution ($512 \times 512$) feature maps.} \\
\caption{Plots of the feature intensity of the selected pixels from different regions over all feature channels, near the \textbf{final} stages:
\textit{Face} region has higher feature magnitude than \textit{background}, while the \textit{artifact} region contain much higher feature magnitude than both \textit{face} and \textit{background} regions.
}
\label{fig:artcause}
\vspace{-0.1in}
\end{figure}

To verify such observation, we plot the amplitudes of the feature at various selected pixels over all feature channels.
Figure~\ref{fig:artcause} (a) compares the plots of the feature intensity at a randomly selected pixel from the background region to that from the face region, as an example. 
In general, the magnitudes of the face region are higher than that of the background region, while they are still in the same order.
On top of the plots in Fig.~\ref{fig:artcause} (a), Fig.~\ref{fig:artcause} (b) further includes the plot at a randomly selected pixel from the artifact region, which is on average one-order higher.
Similar comparison has been observed in higher-resolution intermediate layers, and Fig.~\ref{fig:artcause} (c) shows an example for the $512 \times 512$ feature maps which is next to the final output layer.

To preserve the face region, which is the major component of the image, the trained kernel typically scales to its corresponding magnitude.
Given that the artifact regions have relatively smaller size, but higher magnitudes, they will more likely be propagated or even amplified through all layers, which lead to the prominent circular artifacts in the generated images.
Therefore, the vanilla StyleGAN can hardly remove or suppress any circular artifacts once they are synthesized in any of the intermediate feature maps.


\subsection{Cause of Circular Artifacts}

To systematically investigate the cause of the circular artifacts generated by vanilla StyleGAN, we first raise the following claims followed by the corresponding justifications:
\begin{enumerate}
    \item With high probability (w.h.p.), the intermediate layer of vanilla StyleGAN will generate a feature map containing certain region (i.e. artifact region) with high magnitude, which does not correspond to the facial region.
    \item The artifact regions with high-magnitude pixels will be propagated down the layers of vanilla StyleGAN, w.h.p.
    \item Instance norm (IN) used in vanilla StyleGAN usually amplifies the artifact-region magnitude to be even larger than that of the facial regions.
\end{enumerate}

\begin{figure}[!t]
\centering
\includegraphics[width=8.4cm]{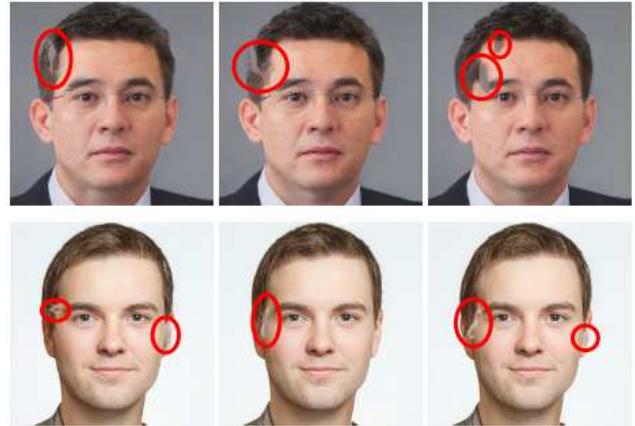}
\caption{Generated images by running the StyleGAN-FFHQ generator on the same latent vector multiple times with random noise inputs. The face shows negligible changes, but the artifact region changes noticeably.}
\label{fig:NoiseArt}
\vspace{-0.1in}
\end{figure}

It is clear that the above three claims comply with our case study and empirical observations. 
Furthermore, we provide detailed justifications of each claim as follows.

\textbf{Claim 1:} \textit{The intermediate layer of vanilla StyleGAN will generate a feature map containing certain region (i.e. artifact region) with high magnitudes, which does not correspond to the facial region, w.h.p..}

In the training stage, vanilla StyleGAN focuses on learning the important face features and prioritizing the accurate synthesis of faces.
Contrastingly, the learned model (e.g. kernel parameters, style modification, etc.) is less correlated with the background regions of the training data.
As the injection of noise in the vanilla StyleGAN introduces random variables that are uncorrelated with the learned model, it leads to randomness in various feature maps with sufficient variance among the non-facial regions (e.g. background) across all channels.
Furthermore, in the vanilla StyleGAN architecture, each layer has a width up to 512 channels, which lead to a high probability that \textit{some} feature map in \textit{one} of these layers has a region of pixels with particularly high magnitudes.

Our experiments (more results and plots are in the supplementary materials) show that in the feature maps at the early stages after the 2nd AdaIN layer, the magnitude of the artifact region are only extremely large in a few channels that are sparsely distributed at random, but small elsewhere. In contrast, the magnitude of the face region relatively smaller variance across all channels~\footnote{Though images are known to be approximately sparse in the transform domain~\cite{wen2015structured}, the magnitudes of coefficients typically follow Laplacian distribution with moderate variance~\cite{dong2012nonlocally}.}.
We conjecture that the location of the high-magnitude artifact channels are initialized by the random noise inputs. To verify our conjecture, we re-run the StyleGAN generator multiple times, fixing the latent vector while applying different random noise inputs.
Figure~\ref{fig:NoiseArt} shows three face images we obtained by only varying the noise inputs.
It is clear that the only major difference~\footnote{We did not observe any major change within the face region due to the different noise input.} is the location of the generated artifacts (highlighted in red circles), which can be attributed solely to the random noise.
The empirical results suggest that the artifacts are NOT from the learned generator, but are triggered by the random noise input instead.



\begin{figure}[!t]
\centering
\includegraphics[width=8.4cm]{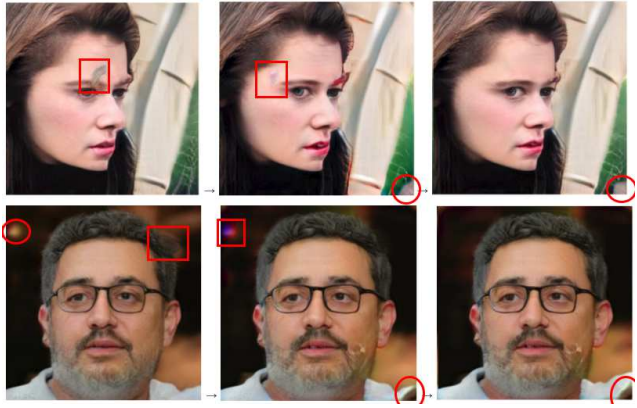}
\caption{Results from ablating units at $8\times8$ resolution: Peak activation at the artifact region is marked by rectangles, while circles highlight the other `section' of the artifact region. 
In both examples, the position of the artifact changes, but the artifact does not entirely disappear.
Further ablation of units has minimal effect on the position of the artifact.}
\label{fig:8x8ablate}
\vspace{-0.1in}
\end{figure}

\textbf{Claim 2:} \textit{The artifact regions with high-magnitude pixels will be propagated down the layers of vanilla StyleGAN, w.h.p..}

In practice, there are much more channels with high-magnitude activation near the \textit{final} stage shown in Fig.~\ref{fig:artcause}, comparing to those in \textit{early} stages.
We conjecture that the randomly initiated sparse artifacts in various feature maps will all be propagated down the layers toward the final output, to form the prominent circular artifacts eventually.
To visualize the artifact propagation, we adapted the GANDissect \cite{GANDissec_2018} framework to StyleGAN trained on faces, to visualize how a specific hidden unit affects the generated images. 
Since there are only a few channels with high-activation magnitudes, we can ablate the convolutional units with maximum activation at the artifact region. 
To further control that \textit{NO} change is caused by random noise, we trained a ``noiseless'' (i.e., without noise input) StyleGAN model when ablating the convolutional units at different resolutions. 

Figure~\ref{fig:8x8ablate} shows the generated images, as an example, by ablating the unit at $8 \times 8$ resolution. We observe that by ablating one single convolutional unit, it is sufficient to alter the location of the generated artifacts, attributing these patterns to even very early stages. Furthermore, by ablating the additional convolutional unit with highest activation at the new location of the artifact, the artifact moves again, which implies that multiple channels may together generate one single artifact. 

\begin{figure}[!t]
\centering
\includegraphics[width=8.4cm]{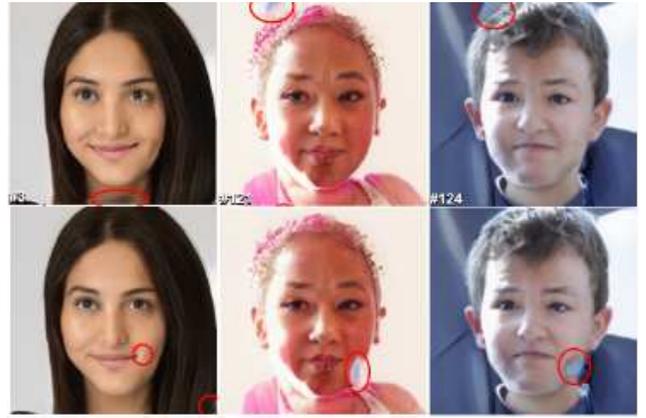}
\caption{Ablating a selected unit at the $8\times8$ layer by using GANDissect for several images. The artifact changes position in all 3 instances when this same unit is ablated.}
\label{fig:8x8Art}
\vspace{-0.1in}
\end{figure}

\begin{figure}[!t]
\centering
\includegraphics[width=8.4cm]{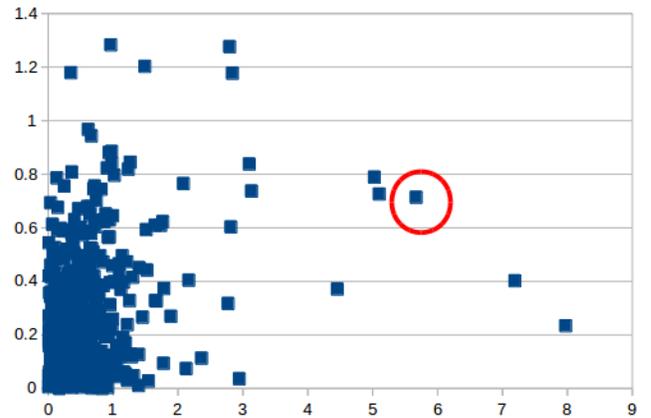}
\caption{Scatter plot of $|\textbf{b}_{\sigma}|$ over $|\textbf{b}_{\mu}|$ at each channel for the 2nd $8\times8$ AdaIN layer. The circled point corresponds to the unit selected in Figure \ref{fig:8x8Art} whose ablation alters the position of the artifact across images.}
\label{fig:BiasPlot}
\end{figure}

Our experimental results show that it is often the same unit with the highest activation at the final artifact region. Figure \ref{fig:8x8Art} shows that ablating this specific unit can alter the location of the final artifact in many images. 
We claim that the artifact is actually determined by the the unit with the largest-magnitude artifact region.
We verify this claim by studying the AdAIN layer which transforms input feature $\textbf{x}$ to $\sigma_y\cdot\frac{\textbf{x}-\mu(\textbf{x})}{\sigma(\textbf{x})}+\mu_y$. 
Here $\mu(\textbf{x})$ and $\sigma(\textbf{x})$ denote the mean and standard deviation of magnitudes of \textbf{x}, respectively. 
Besides, by applying the trained affine transforms to the intermediate latent vector $\textbf{w}$, we have $\mu_y = \textbf{v}_{\mu}\cdot\textbf{w}+\textbf{b}_{\mu}$ and $\sigma_y = \textbf{v}_{\sigma}\cdot \textbf{w}+\textbf{b}_{\sigma}$.
Figure~\ref{fig:BiasPlot} shows the scatter plot of $|\textbf{b}_{\sigma}|$ over $|\textbf{b}_{\mu}|$ for the 2nd $8\times8$ AdaIN layer. 
The unit highlighted in Fig.~\ref{fig:8x8Art} turns out to have large $|\textbf{b}_{\sigma}|$ and $|\textbf{b}_{\mu}|$, which lead to large $\mathbb{E}(|\mu_y|)$ and $\mathbb{E}(|\sigma_y|)$, respectively. Since $|\mu_y|$ and $|\sigma_y|$ denotes the magnitude of the translation and scaling factor respectively after IN, it is very likely to generate high-magnitude artifacts at the corresponding region.

\begin{figure}[!t]
\centering
\includegraphics[width=8.4cm]{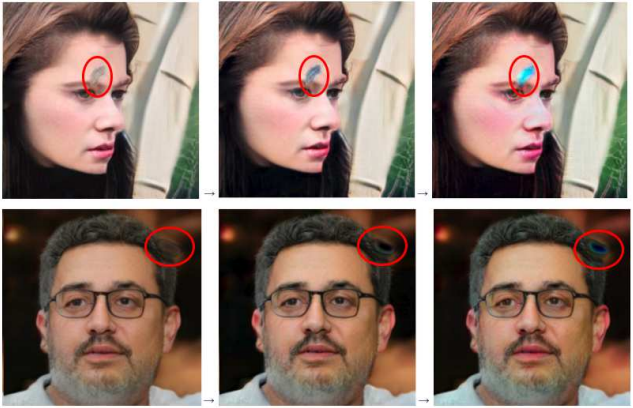}
\caption{Ablating one unit at a time in the `artifact' region by selecting the unit in the $128\times128$ layer with highest activation iteratively: Only the colour or texture of the artifact region changes, while the location of the artifacts is preserved.}
\label{fig:128x128ablate}
\vspace{-0.1in}
\end{figure}

\begin{figure}[!t]
\centering
\includegraphics[width=8.4cm]{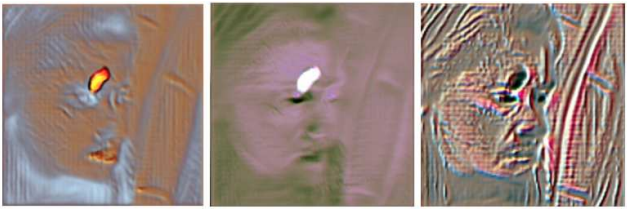}
\caption{Results by ablating all but one unit in the $128\times128$ layer. The location of the artifact region is consistent in all generated images.}
\label{fig:128x128unit}
\vspace{-0.1in}
\end{figure}

Figure~\ref{fig:128x128ablate} shows the results by iteratively ablating one unit at the $128 \times 128$ layer, with highest activation at the artifact region.
Different from the results by ablating the \textit{early-stage} units in Fig.~\ref{fig:8x8ablate}, ablating each unit at a time affects more regional features, e.g., the colour or texture of the artifact region, rather than the location of the artifacts.
Similarly, Fig.~\ref{fig:128x128unit} shows the results by ablating all but one unit in the $128\times128$ layer, where the location of the artifacts is still preserved.
It shows that the most prominent artifact ``spreads'' to more units deeper in the generator, where the corresponding intensity depends primarily on the magnitude of $\mu_y$ and $\sigma_y$ at each unit. More explanations with toy examples are presented in the supplementary materials.




\textbf{Claim 3:} \textit{Instance norm (IN) used in vanilla StyleGAN usually amplifies the artifact-region magnitude to be even larger than that of the facial regions.}

Claim 1 and Claim 2 suggest that some high-magnitude artifacts will randomly emerge in the early or intermediate feature maps, which are likely to be propagated down the layers. 
The remaining question is how these artifacts are amplified and become so significant in the final output. 

\begin{figure}
    \centering
        \includegraphics[width=4cm]{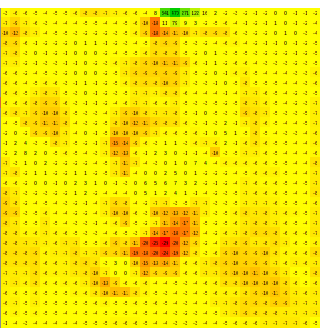}
        \includegraphics[width=4cm]{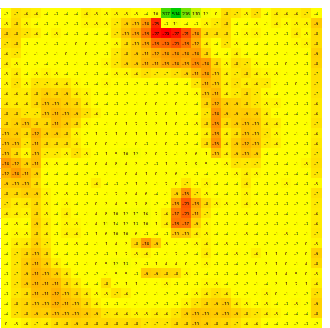}
        \includegraphics[width=4cm]{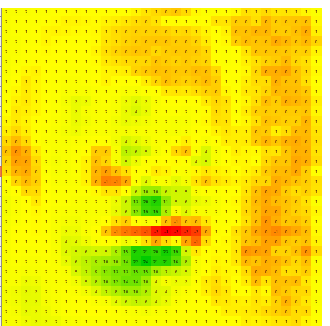}
        \includegraphics[width=4cm]{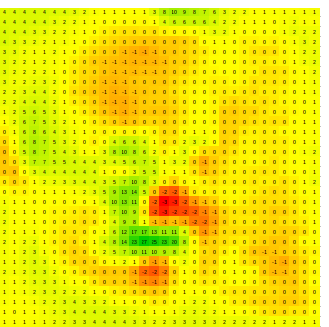}
    \caption{Feature Maps at 32x32 resolution. \\ Top: Artifact-containing maps, where the high activation region is an artifact that only occupies 3-4 pixels. \\ Bottom: Non artifact-containing maps, where the high activation regions approximately correspond to facial features.} 
\label{fig:32Art}
\vspace{-0.1in}
\end{figure}

We provide some analysis showing that IN is responsible for the artifact amplification:
Consider an $l\times l$ feature map generated by an intermediate layer, and we take the magnitude (i.e., absolute value) of the pixel values.
Denote the set of $\alpha l^2$ pixels with highest magnitudes as $\mathbf{S}_1$ with $\alpha \le 0.5$. Correspondingly, the rest low-magnitude $(1-\alpha)l^2$ pixels as $\mathbf{S}_2$.
Denote the mean and variance of the magnitudes of pixels in $\mathbf{S}_1$ and $\mathbf{S}_2$ as $\mu_1 > 0$, $\sigma_1^2$, $\mu_2 > 0$, and $\sigma_2^2$, respectively. 
Thus, the overall mean and variance of the $l^2$ pixels are 
\begin{align}
\nonumber \mu = & \; \alpha \mu_1 + (1-\alpha) \mu_2 \\
\sigma^2 = & \; \sigma_1^2 \alpha+ \sigma_2^2 (1-\alpha) + \alpha(1-\alpha)(\mu_1-\mu_2)^2\;\;.
\end{align}
After IN, the mean magnitude of all pixels in $\mathbf{S}_1$ becomes
\begin{align}
\nonumber \hat{\mu}_1 = & \frac{\mu_1 - \mu}{\sigma}  \\
\nonumber = & \frac{\mu_1-(\alpha \mu_1 + (1-\alpha)\mu_2)}{\sqrt{\sigma_1^2 \alpha + \sigma_2^2 (1-\alpha) + \alpha(1-\alpha)(\mu_1-\mu_2)^2}} \\
\nonumber = & \frac{(1-\alpha)(\mu_1-\mu_2)}{\sqrt{\sigma_1^2 \alpha + \sigma_2^2 (1-\alpha) + \alpha(1-\alpha)(\mu_1-\mu_2)^2}} \\
\approx & \frac{(1-\alpha)\cdot \mu_1}{\sqrt{\alpha(1-\alpha)\mu_1^2}} = \sqrt\frac{1-\alpha}{\alpha} \;\; ,
\end{align}
assuming that $|\mu_2|<<|\mu_1|$ and $\sigma_1, \sigma_2 <<|\mu_1|$. Here $\hat{\mu}$ is monotonically decreasing as a function of $\alpha$, thus $\hat{\mu}_1$ will be significantly amplified when the high-magnitude region is minority, i.e., $\alpha$ is small, which is mostly true in our extensive experiments. 
As shown in the results of GANDissect, the high-magnitude region corresponds to the generation of the particular feature in that region. At low-mid resolutions, these involve entire features (e.g. eyes/nose/mouth). In contrast, artifacts are typically initiated at random which are sparsely distributed (e.g., only several pixels in the feature maps). 
Figure \ref{fig:32Art} shows the comparison of high-magnitude pixel distribution in artifact and face region, by visualizing the $32 \times 32$ feature maps.

In summary, the combination of the Claims 1 to 3 explains how the circular artifacts are synthesized, propagated, and amplified in generated images by the vanilla StyleGAN. 


\subsection{Characteristics of Artifacts}
\begin{itemize}
\item \textbf{Why are the artifacts circular?}
This can be explained by the $3\times3$ convolutional kernel.  Pixels at the center of the generated artifact are more likely to have high magnitudes neighbours, whereas those at the sides only have 2-3 high-magnitude neighbours. This means that the probability of continuing to have a high magnitude is lower at the sides of the artifact. As a result, the magnitude is highest at the center, and decreases towards the boundaries. In addition, the location of the high-magnitude artifact regions across various feature maps are similar, but not perfectly aligned. The final image is generated by weighted combination of several artifact regions from different feature maps, each of which has a slightly different patterns. Thus, it leads to the ring-shaped boundary of the artifacts in the final layer, which has lower magnitude than the interior part of the artifact forming the circular structures eventually.

\item \textbf{Why aren't there many artifacts?}
As discussed in the Claim 2, the strongest artifact will persist. However, we have yet to explain why the other artifacts do not also persist, leading to a generated face that is full of artifacts. Indeed there are some images where the artifact can split into 2 parts - but these are relatively rare. From Fig.~\ref{fig:NoiseArt} to \ref{fig:8x8Art}, it is clear that the actual size of the artifact region does not change much. 
For the spreading described in Claim 2 to occur, the artifact region must be sufficiently large. On the other hand, for the activation magnitudes to escalate as described in Claim 3, the artifact region must be sparse. Therefore, only those artifacts within a range of sizes can persist. As the total size of the artifacts are bounded, it limits the number of artifacts in a single generated image.



\end{itemize}

\section{Proposed Solution for Artifact Removal}
Inspired by the recent idea of batch-instance normalization (BIN) \cite{BIN_2018}, we propose a novel pixel-instance normalization (PIN) approach, which balances between IN and pixel normalization (PN). The proposed PIN will be used as the normalization layer in the improved StyleGAN to remove the undesired circular artifacts.

Denote the input image as $\textbf{x} \in \mathbb{R}^{C\times H \times W}$, where $C,H,W$ are the number of channels, the image height and width, respectively. 
The scalar $x_{chw}$ denotes the $chw$-th element of $\textbf{x}$, where $c$ is the channel index, and $h,w$ are the spacial coordinates.
The PN operation $\eta^P ( \cdot )$ is defined as $\textbf{y}^P = \eta^P (\textbf{x})$, where
\begin{align}
y^P_{chw} \; = \; x_{chw} \; / \; \sqrt{\frac{1}{C} \displaystyle \sum_{i = 1}^C x_{ihw}^2 + \epsilon}  \;\;\;\; \forall c,h,w 
\end{align}

In vanilla StyleGAN, IN operator, denoted as $\eta^I ( \cdot )$, is used as the normalization layer.   
The IN operation $\textbf{y}^I = \eta^I (\textbf{x})$ is defined as 
\begin{align}
y_{chw}^{I} = \frac{x_{chw}-\mu_{c}^{(I)}}{\sqrt{\sigma_{c}^{2(I)}+\epsilon}}
\end{align}
where the mean and standard deviation are defined as
\begin{align}
\mu_{c}^{(I)} = & \frac{1}{HW} \displaystyle \sum_{h=1}^H \displaystyle \sum_{w=1}^W x_{chw} \\
\sigma_c^{2(I)} = & \frac{1}{HW} \displaystyle \sum_{h=1}^H \displaystyle \sum_{w=1}^W \left(x_{chw}-\mu_c^{(I)}\right)^2
\end{align}

We propose the PIN operator, i.e., $\textbf{y} = \eta^{PI} (\textbf{x})$, which acts as the weighted combination of PN and IN as
\begin{align}
    \textbf{y} \; = \; \rho \, \cdot \, \textbf{y}^P + \, (1-\rho) \, \cdot \, \textbf{y}^I\;\;,
\end{align}
where $\rho \in \left[0,1\right]^C$ are trainable, and $\cdot$ denotes the channel-wise product.
The subsequent style modification layer is retained as an affine transformation. 
With the proposed modification, the output will be $\textbf{y'} = \gamma \cdot \textbf{y}  + \beta$, where $\gamma, \beta \in \mathbb{R}^C$ are learnable parameters as before.
We constrained the components of $\rho$ to be $0 \leq \rho_i \leq 1$ by clipping. 
The proposed PIN is a generalized fusion of PN and IN, e.g., when $\rho = \textbf{0}$ or $\textbf{1}$, PIN reduces to IN or PN, respectively.

\section{Experiments}

\begin{figure}[!t]
\centering
\includegraphics[width=8.4cm]{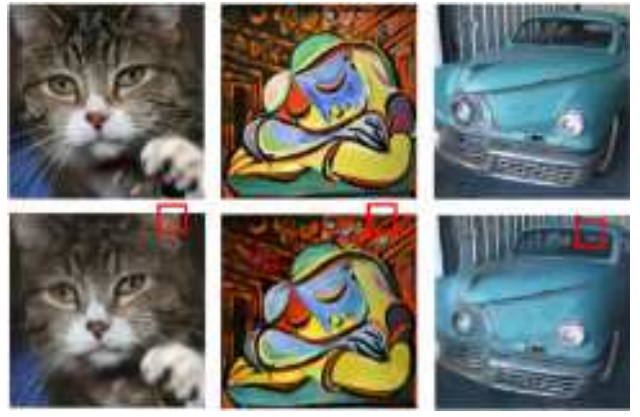}
\caption{Figure taken from \cite{Image2Style_2019}. The top row contains the target image, while the bottom row is the embedded image. The circular artifacts were manually identified.}
\label{fig:EmbedArt2}
\vspace{-0.1in}
\end{figure}

\subsection{Ablation Study for Vanilla StyleGAN}

We present an ablation study to verify our Claim 3, i.e., IN used in vanilla StyleGAN is responsible for the circular artifacts. 
First, the circular artifacts are unlikely caused by insufficient training of the StyleGAN model or hyperparameter tuning, as the artifacts also appear in the curated examples provided by the authors of StyleGAN~\footnote{ https://drive.google.com/drive/folders/1-l46akONUWF6LCpDoeq63H53rD7MeiTd}.
Furthermore, Fig.~\ref{fig:EmbedArt2} shows that similar circular artifacts continued to appear in the Image2StyleGAN embedding of non-face objects such as cars, paintings and cats in an expanded latent space $\mathcal{W}^{18}=\mathbb{R}^{512\times18}$. 
This suggests that the network inherently produces artifacts irrespective of latent vector $\textbf{z}$ or $\textbf{w}$.

The precursor to StyleGAN was the PRGAN generator \cite{PRGAN_2017}, which starts with low resolution $4\times4$ images and progressively increases the resolution to $8\times8$, $16\times16$, etc., i.e., the architecture in Fig.~\ref{fig:StyleGANarch} (a).
The StyleGAN architecture retains this progressive structure with 2 layers per resolution, while only modifying the network architecture in Fig.~\ref{fig:StyleGANarch} (b).
However, the circular artifacts only appear in the StyleGAN-generated images, whereas not being the major issue for PRGAN. 
Therefore, we performed ablation testing by reverting each of major changes introduced in Fig.~\ref{fig:StyleGANarch} (b). 

Specifically, we trained the generator over the Flickr-Faces-HQ Dataset (FFHQ), dubbed StyleGAN-FFHQ~\cite{StyleGAN_2018}, with the modifications listed in Table \ref{tab:ablation}, to identify which major component introduced to the StyleGAN model comparing to PRGAN is responsible for generating the circular artifacts. 
Here, we separate the AdaIN \cite{styleTransfer_2017} into 2 separate steps for ablation test - an Instance Normalization (IN) \cite{IN_2016} followed by the Style Modification, i.e. an affine transform. 

\begin{figure}
    \centering
        \includegraphics[width=2.6cm]{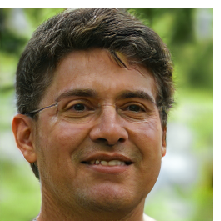}
        \includegraphics[width=2.6cm]{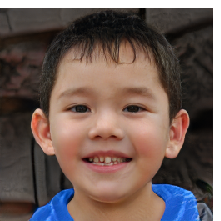}
        \includegraphics[width=2.6cm]{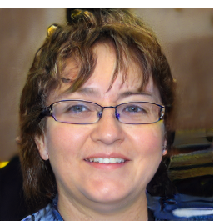}
    \caption{Sample images generated by StyleGAN-FFHQ with PN instead of IN}
    \label{fig:PNorm}
    \vspace{-0.1in}
\end{figure}

The results show that only by replacing the PN with IN in StyleGAN can we remove the circular artifacts in the generated images. 
Figure~\ref{fig:PNorm} includes some results using the modified StyleGAN with PN. 
Though the generated faces no longer contain circular artifacts, the quality of the generated image degraded compared to those by vanilla StyleGAN.

\begin{table}
\begin{tabular}{ | l | l | }
\hline
\textbf{Test} & \textbf{Circular Artifacts?} \\
\hline
Use CelebA dataset & Yes \\
\hline
Remove mapping network & Yes \\
\hline
Use traditional input & Yes \\
\hline
Remove noise & Yes \\
\hline
Remove blurring & Yes \\
\hline
Remove style transfer & Yes \\
\hline
Use pixel normalization (PN) & No \\
\hline
\end{tabular}
\caption{Ablation Study: Reverting each of the major changes evolving from PRGAN to StyleGAN.}
\label{tab:ablation}
\end{table}

\subsection{Circular Artifacts Removal by PIN}

\begin{figure}
    \centering
        \includegraphics[width=2.6cm]{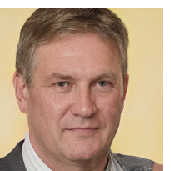}
        \includegraphics[width=2.6cm]{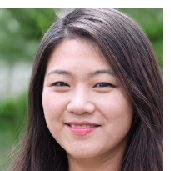}
        \includegraphics[width=2.6cm]{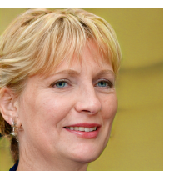}
    \caption{Sample images generated by StyleGAN-FFHQ with PIN}
    \label{fig:PIN}
    \vspace{-0.1in}
\end{figure}

We investigate the effectiveness of the proposed PIN by applying it to StyleGAN for face image generation.
All the combination weights are initialized as $\rho = \textbf{0}$, i.e., initialized as the original StyleGAN configuration, for training the StyleGAN-FFHQ with PIN.
Figure~\ref{fig:PIN} shows several generated images using the improved model, and they all demonstrate better visual quality, comparing to those using PN in Fig.~\ref{fig:PNorm}, without any circular artifacts. 
We include more examples by the trained StyleGAN-FFHQ with PIN in the supplementary materials, in which we observe similar visual quality improvement in all generated images.

\begin{figure}[h]
\centering
\includegraphics[width=8.4cm]{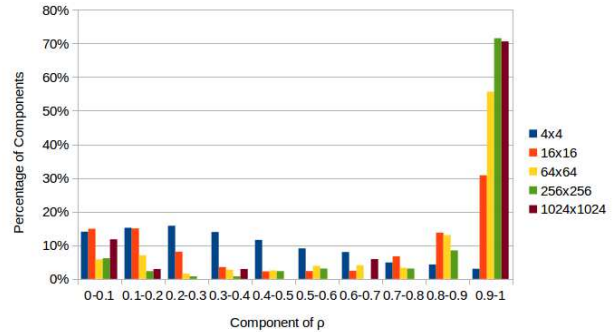}
\caption{Histogram of components of $\rho$ for selected resolutions. $0$ corresponds to only IN, while $1$ corresponds to only PN. The earlier layers tend to have more components using IN, while the later layers largely use PN.}
\label{fig:GateFreq}
\vspace{-0.1in}
\end{figure}

Figure~\ref{fig:GateFreq} visualizes the learned $\rho$'s at each layer using a histogram. 
It is obvious that the learned PINs did not directly reduce to IN or PN. Instead, IN is more likely to dominate in the earlier layers, while PN is more preferred in the later layers. 
Our conjecture is that applying IN in early stages helps the generator to quickly transfer desired styles into face features. 
By applying PN subsequently, the activation magnitudes will not escalate to form artifact regions.


Although we have demonstrated that PIN is a simple method for circular artifact removal in StyleGAN, it is not the only approach to tackle this challenge. For example, one can also apply the recently introduced switchable normalization~\cite{Switchable_2019}, with PN as one of the normalizations. More advanced solution has also been proposed in improved version of StyleGAN \cite{style-gan2}, in which AdaIN layer is redesigned and replaced with a normalization technique called weight demodulation. We leave these possibilities for future work.


\section{Conclusion}
In this paper, we have highlighted the recurring problem of circular artifacts in StyleGAN-generated faces, and found that it is due to the selective escalation of activation magnitudes, promoted by the StyleGAN architecture. Specifically, the effects of IN are compounded through the progressive structure of StyleGAN, resulting in these artifacts that are neither seen in stand-alone PRGAN or AdaIN style transfer. Additionally, we have proposed a simple solution that can remove the circular artifacts without excessive loss of image quality. Though \cite{style-gan2} had proposed an advanced solution to deal with artifacts for StyleGAN, we believe that the systematic analysis approach in this study should be useful for the exploration of the functionalities of various GAN architectures, especially can be used to explain the generation of other unusual artifacts. 


\bibliography{ref.bib}
\bibliographystyle{aaai}
\end{document}